\documentclass{article}

\usepackage{microtype}
\usepackage{graphicx}
\usepackage{subcaption}
\usepackage{booktabs}

\def\method{DREAM}

\usepackage{hyperref}

\usepackage[preprint]{icml2026}

\usepackage{amsmath}
\usepackage{amssymb}
\usepackage{mathtools}
\usepackage{amsthm}

\usepackage[capitalize,noabbrev]{cleveref}

\theoremstyle{plain}
\newtheorem{theorem}{Theorem}[section]

\theoremstyle{definition}

\newtheorem{assumption}[theorem]{Assumption}
\theoremstyle{remark}
\newtheorem{remark}[theorem]{Remark}

\usepackage[textsize=tiny]{todonotes}

\def \x {\mathbf{x}}
\usepackage{bm}
\usepackage{amsfonts, amssymb, amsmath}
\usepackage{booktabs}
\usepackage{multirow}
\usepackage{makecell}
\usepackage{subcaption}
\usepackage{siunitx}
\usepackage{xcolor}
\usepackage{colortbl}
\usepackage{siunitx}
\usepackage{times}
\usepackage{helvet}
\usepackage{courier}
\usepackage{algorithm}
\usepackage{algorithmic}
\usepackage{enumitem}
\DeclareMathOperator*{\argmax}{argmax}
\definecolor{tableheadcolor}{RGB}{227, 233, 251}
\definecolor{tablerowcolor}{RGB}{242, 242, 242}
\definecolor{ourrowname}{RGB}{241, 223, 221}

\icmltitlerunning{DREAM: Dual-Standard Semantic Homogeneity with Dynamic Optimization for Graph Learning with Label Noise}

\begin{document}

\twocolumn[
  \icmltitle{DREAM: Dual-Standard Semantic Homogeneity with Dynamic Optimization for Graph Learning with Label Noise}



  \icmlsetsymbol{equal}{*}

  \begin{icmlauthorlist}
    \icmlauthor{Yusheng Zhao}{equal,pku}
    \icmlauthor{Jiaye Xie}{equal,pku}
    \icmlauthor{Qixin Zhang}{equal,ntu}
    \icmlauthor{Weizhi Zhang}{uic}
    \icmlauthor{Xiao Luo}{wisc}
    \icmlauthor{Zhiping Xiao}{uw}
    \icmlauthor{Philip S. Yu}{uic}
    \icmlauthor{Ming Zhang}{pku}
  \end{icmlauthorlist}

  \icmlaffiliation{pku}{National Key Laboratory for Multimedia Information Processing, School of Computer Science, Peking University-Anker Embodied AI Lab, Peking University}
  \icmlaffiliation{ntu}{College of Computing and Data Science, Nanyang Technological University}
  \icmlaffiliation{uic}{Department of Computer Science, University of Illinois Chicago}
  \icmlaffiliation{uw}{Paul G. Allen School of Computer Science and Engineering, University of Washington}
  \icmlaffiliation{wisc}{Department of Statistics, University of Wisconsin–Madison}

  \icmlcorrespondingauthor{Firstname1 Lastname1}{first1.last1@xxx.edu}
  \icmlcorrespondingauthor{Firstname2 Lastname2}{first2.last2@www.uk}

  \icmlkeywords{Machine Learning, ICML}

  \vskip 0.3in
]



\printAffiliationsAndNotice{}  

\begin{abstract}
Graph neural networks (GNNs) have been widely used in various graph machine learning scenarios. Existing literature primarily assumes well-annotated training graphs, while the reliability of labels is not guaranteed in real-world scenarios.
Recently, efforts have been made to address the problem of graph learning with label noise. However, existing methods often \textit{(i)} struggle to distinguish between reliable and unreliable nodes, and \textit{(ii)} overlook the relational information embedded in the graph topology. 
To tackle this problem, this paper proposes a novel method, \underline{D}ual-Standa\underline{r}d S\underline{e}mantic Homogeneity with Dyn\underline{a}mic Opti\underline{m}ization (\method{}), for reliable, relation-informed optimization on graphs with label noise.
Specifically, we design a relation-informed dynamic optimization framework that iteratively reevaluates the reliability of each labeled node in the graph during the optimization process according to the relation of the target node and other nodes. To measure this relation comprehensively, we propose a dual-standard selection strategy that selects a set of anchor nodes based on both node proximity and graph topology. Subsequently, we compute the semantic homogeneity between the target node and the anchor nodes, which serves as guidance for optimization. We also provide a rigorous theoretical analysis to justify the design of \method{}.
Extensive experiments are performed on six graph datasets across various domains under three types of graph label noise against competing baselines, and the results demonstrate the effectiveness of the proposed \method{}.
\end{abstract}

\begin{figure}
    \centering
    \includegraphics[width=\linewidth]{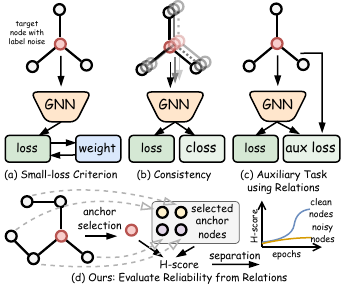}
    \vspace{-4mm}
    \caption{Conventional methods either overlook the relational information in distinguishing noisy labels (\emph{e.g.}, small loss criterion), or resort to regularization for extra supervision signals that are hard to separate clean from noise nodes (\emph{e.g.}, consistency regularization, auxiliary tasks using edges). By comparison, this paper explicitly evaluates the reliability of nodes from carefully selected relational information in the graph topology.}
    \label{fig:motivation}
    \vspace{-5mm}
\end{figure}
\section{Introduction}
Graph neural networks (GNNs) \cite{kipf2016semi, li2018deeper, wu2019simplifying} have become the primary paradigm for learning representations on non-Euclidean graph-structured data, enabling a plethora of applications in various fields, including recommendation systems \cite{wu2019session, yang2021consisrec, wang2024distributionally, ou2025ls}, social network analysis \cite{zhang2022improving, yin20222, jain2023opinion, li2023survey}, spatio-temporal forecasting \cite{wang2022causalgnn, lan2022dstagnn, hu2025lightst}, and protein-protein interactions \cite{shen2021npi, jha2022prediction, reau2023deeprank, shuvo2025equirank}. 
Despite their superior performance, most research in this field assumes that graphs are well-annotated, whereas in reality, annotation noise is inevitable and harms model performance \cite{frenay2013classification, johnson2022survey, wang2024noisygl}. 

Recently, the task of graph learning with label noise has received increasing attention, with a number of methods proposed \cite{dai2021nrgnn, dai2022towards, yin2023omg, zhu2024robust, wu2025soft}. 
One line of research focuses on regularizing the loss function through information from the graph structure \cite{dai2021nrgnn, du2021noise, li2024contrastive, wu2024mitigating} or node semantics \cite{lu2024noise, zhu2024robust}. 
Another line of research reweights the samples according to their reliability, typically measured by the loss function \cite{gui2021towards, wei2022self, li2024regroup, liao2025instance}, where a smaller loss indicates reliability of the sample \cite{yu2019does, kim2021fine}.

However, prior research often suffers from two major shortcomings. 
\textit{(i) Existing works frequently struggle to differentiate between reliable and unreliable nodes within complex graph topology}. When faced with potential label noise, many approaches resort to auxiliary supervision signals, such as contrastive learning \cite{xue2022investigating, zhu2024robust} or regularization \cite{du2021noise, dai2021nrgnn}, while either falling short in or altogether avoiding the task of distinguishing reliable and unreliable nodes in the complex graph topology. Explicit identification of these nodes can not only mitigate the impact of noisy labels but also prioritize the importance of nodes, benefiting model optimization \cite{hu2020graph, zhang2022galaxy, yuan2024graph}.
\textit{(ii) Prior works often neglect the rich relational information naturally embedded in the graph topology}. Many works reweight the importance of samples following the small-loss criterion \cite{liu2015classification, gui2021towards, dai2022towards, chen2024learning}. However, they often ignore the relational information embedded in the graph topology, which is beneficial to identifying the reliability and informativeness of labeled nodes \cite{chen2021weak, kim2023neural}.

To address these limitations, this paper proposes a novel method named \underline{D}ual-Standa\underline{r}d S\underline{e}mantic Homogeneity with Dyn\underline{a}mic Opti\underline{m}ization (\method{}) for reliable and relation-informed dynamic optimization of graph machine learning models in the presence of label noise.
As illustrated in Figure \ref{fig:motivation}, the core idea of \method{} is to leverage the rich relational information within the graph to reevaluate the reliability of each labeled node in the graph iteratively during optimization.
To achieve this goal, we design dual-standard semantic homogeneity that measures the homogeneity between each target node and a set of carefully selected anchor nodes.
Specifically, \method{} adopts both proximity-aware and topology-aware anchor selection to identify relevant candidate nodes that form the anchor set for each target node. Subsequently, the homogeneity score between the target node and its anchors is computed based on the similarity of their latent representations obtained during the forward pass. Finally, these homogeneity scores are used as indicators to reweight the importance of each labeled node during optimization. This process is integrated into the model's supervision and performed iteratively, enabling relation-informed dynamic optimization.
We also provide a rigorous theoretical analysis of \method{}, showing that an $\epsilon$-error approximation to the optimal parameters is guaranteed under the proposed training framework.
We perform extensive experiments across six graph datasets from different domains under three different types of label noise. The results compared with a number of baselines validate the superiority of the proposed \method{}.

The contribution of this paper is summarized as follows:
\begin{itemize}[leftmargin=3mm, topsep=1mm, itemsep=1mm]
    \item \textbf{New Perspective}: We propose to utilize the rich relational information naturally embedded in the graph topology to dynamically re-evaluate the reliability of each labeled node during optimization.
    \item \textbf{Novel Methodology}: We design a novel method, termed dual-standard semantic homogeneity with dynamic optimization (\method{}), which measures the reliability of each labeled node through a set of node-specific anchors selected based on both proximity and graph topology.
    \item \textbf{Theoretical Insights}: We provide a rigorous theoretical analysis, showing that the proposed \method{} achieves an $\epsilon$-error approximation to the optimal model parameters.
    \item \textbf{Extensive Experiments}: We perform extensive experiments across various datasets and different types of label noise against competing baselines, and the results confirm the effectiveness of the proposed \method{}.
\end{itemize}
\begin{figure*}
    \centering
    \includegraphics[width=\linewidth]{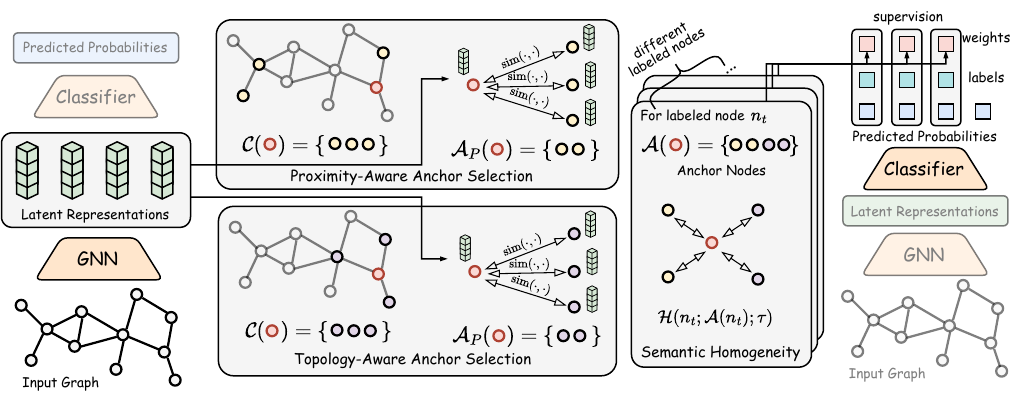}
    \vspace{-3mm}
    \caption{The overall framework of the proposed \method{}. \method{} first selects a set of anchors relative to each target node using proximity-aware anchor selection and topology-aware anchor selection. Then, the semantic homogeneity score is computed according to the relation between the anchors and the target node. Finally, the semantic homogeneity score is used as an indicator of the reliability of each labeled node to reweight the optimization objective dynamically during optimization. }
    \vspace{-2mm}
    \label{fig:framework}
\end{figure*}

\section{Related Works}
\subsection{Graph Neural Networks}
Graph neural networks (GNNs) \cite{kipf2016semi, hamilton2017inductive, velivckovic2017graph, xu2018powerful} have been extensively applied in many graph machine learning tasks \cite{wu2020comprehensive, corso2024graph}, facilitating downstream applications in various fields including recommendation \cite{wu2019session, ou2025ls}, healthcare \cite{li2020graph, paul2024systematic}, transportation \cite{wang2020traffic, akpaku2025ragn}, and bioinformatics \cite{wang2021scgnn, koh2024physicochemical, airas2025scaling}. The graph neural networks obtain the node embeddings by propagating the representations from adjacent nodes iteratively \cite{xu2018powerful}. Subsequently, the obtained features are used in downstream tasks like classification \cite{kipf2016semi} or regression \cite{jia2020residual, zhang2024regexplainer}. However, most of the models are trained on well-annotated datasets \cite{hu2020open, morris2020tudataset, dwivedi2023benchmarking}, whereas in practice, label noise often exists and hurts the optimization of GNNs \cite{wang2024noisygl, zhu2024robust, han2025uncertainty}. Therefore, this paper studies the problem of graph learning with label noise and proposes a relation-informed dynamic optimization framework to train GNNs robustly against noisy labels.

\subsection{Learning with Label Noise}
Label noise is a widespread problem in deep learning, where part of the training data is incorrectly labeled \cite{frenay2013classification, patrini2017making, lukasik2020does}. In practice, many datasets are manually labeled, leading to potential misclassifications of instances \cite{ekambaram2017finding, yun2021re}. This would hurt the optimization of deep learning models. A number of methods have been proposed that focus on Euclidean data like images or texts \cite{karim2022unicon, penso2024confidence, jiang2024more}. These works often tackle the problem by loss correction \cite{patrini2017making, arazo2019unsupervised, toner2023label} or relabeling them for training \cite{zheng2021meta, huang2022uncertainty, park2023robust, li2024feddiv}. 

Nevertheless, the label noise problem in graph learning is more complex due to the non-Euclidean structure. The label noise on graphs may be uniformly distributed or structure-dependent \cite{wang2024noisygl}. To address the problem, one line of research regularizes the loss objective with additional supervision signals from self-consistency \cite{li2024contrastive, lu2024noise, yin2024sport} or the graph structure \cite{dai2021nrgnn, wu2024mitigating, cheng2024resurrecting}. However, they often struggle to differentiate the noisy labels, hurting the performance when the noise increases. Another line of research reweights the nodes according to certain measurements like the loss function \cite{dai2022towards, yang2023interactive, chen2024erase}. However, they often neglect the rich relational information in the graph topology. While there are efforts on correcting the labels, they often rely on assumptions about the graph structure or patterns of mislabeling \cite{li2022devil, yin2023omg}. In this paper, we propose a novel method named \method{} that utilizes the relational information in the graph topology to dynamically reevaluate the reliability of labeled nodes during optimization for robust graph learning with label noise.
\section{Methodology}
\subsection{Problem Definition}
We denote a graph as $\mathcal G=\langle \mathcal V, \mathcal E, \bm X \rangle$, where $\mathcal V$ is the set of nodes, $\mathcal E$ is the set of edges, and $\bm X$ is the input features of nodes. We assume $\bm X\in\mathbb R^{N\times d_{in}}$, where $N=|\mathcal V|$ and $d_{in}$ is the dimension of the input node feature. To obtain the node embeddings, a graph neural network $f_\theta$ is used, and the learned representations of each node are denoted as $f_\theta(\mathcal G)=\bm Z = [\bm z_1, \bm z_2, \dots, \bm z_N]^\top\in \mathbb R^{N\times d}$, where $d$ is the hidden dimension. Subsequently, a classifier is $g_\phi$ used to obtain the class probabilities $\bm p_i$ of each node. We denote the number of classes as $C$, the set of nodes that have labels as $\mathcal S$, and the noisy labels as $\mathcal Y = \{y_i \mid i\in \mathcal S\}$. The goal is to optimize the model $\mathcal M=f_\theta\circ g_\phi$ with the noisy labels $\mathcal Y$ to achieve better performance on the test nodes.

\subsection{Framework Overview}
We now present the proposed Dual-Standard Semantic Homogeneity with Dynamic Optimization (\method{}), as illustrated in Figure \ref{fig:framework}. To measure the reliability of each labeled node (the target node, denoted as $n_t$), \method{} first selects a set of relevant nodes (the anchors, denoted as $\mathcal A(n_t)\subset \mathcal V$) for each target node $n_t$. Specifically, we design proximity-aware and topology-aware anchor selection mechanisms to construct the anchor set $\mathcal A(n_t)$. Subsequently, we compute a semantic homogeneity score based on the similarity between the target node and its anchors. This score serves as an estimate of the labeled node's reliability and is used to reweight the optimization objective. During the optimization process, we iteratively reevaluate the reliability of each labeled node using the dual-standard semantic homogeneity, enabling robust dynamic optimization.

\subsection{Proximity-Aware Anchor Selection}
We first select the anchor nodes according to the semantic proximity of nodes. For a target node $n_t$ that is correctly labeled as $y_t$, we expect it to have many semantically proximal nodes that are also labeled as $y_t$, and vice versa. By examining the relationship of the target node and its selected proximal nodes, we can assess the reliability of the target node. Therefore, we propose proximity-aware anchor selection to select these nodes as anchors.

Specifically, given target node $n_t$ labeled as $y_t$, we first construct the set of candidate nodes that share the label of $n_t$. Formally, we have the candidate set for $n_t$ as:
\begin{equation}\label{eq:candidate-p}
\mathcal C_P(n_t) = \{n_i\mid i \in \mathcal S \land y_i=y_t\}.
\end{equation}
We then select the semantically proximal nodes from $\mathcal C_P(n_t)$ as proximity-aware anchors. To achieve this, we utilize the learned representations of nodes obtained during the model's forward pass, \emph{i.e.}, $\bm z_i, i\in \mathcal C_P(n_t)$, and a similarity measurement $\operatorname{sim}(\cdot, \cdot)$. Formally, we have:
\begin{equation}
\label{eq:proximity-anchors}
\mathcal A_P(n_t)
=\argmax_{\substack{\mathcal I\subseteq\mathcal C_P(n_t), |\mathcal I|=k_P}}
\sum_{i\in \mathcal I}\operatorname{sim}(\bm z_t,\bm z_i),
\end{equation}
where $k_P$ is the number of proximity-aware anchors. Conceivably, for $n_t$ correctly labeled as $y_t$, the set of $\mathcal C_P(n_t)$ contains a number of nodes similar to $n_t$, as most of them fall into the correct category of $y_t$. Conversely, when $n_t$ is incorrectly labeled, only a few nodes in $\mathcal C_P(n_t)$ fall into the correct category $\hat y_t$ of $n_t$. This observation serves as guidance for evaluating the reliability of the labeled nodes.

\subsection{Topology-Aware Anchor Selection}
In addition to semantic proximity, the graph topology also plays a significant role. For a target node $n_t$ labeled as $y_t$, the category of $y_t$ may contain several sub-categories scattered across the graph. Nodes closer in the topology are more likely to belong to the same sub-category. The goal of topology-aware anchor selection is to identify anchors that potentially belong to the same sub-category as a correctly labeled $n_t$. By comparing the anchors with the target node, we can distinguish reliable nodes from unreliable ones.

Concretely, given two nodes $n_i$ and $n_j$, we measure their topological proximity using the geodesic distance \cite{bondy2008graph}, denoted as $d_\mathcal G(n_i,n_j)$. We then construct the candidate set for $n_t$ according to the topological proximity as follows:
\begin{equation}\label{eq:candidate-t}
\mathcal C_T(n_t) =
\{n_i\mid d_\mathcal G(n_i,n_t)\le d_{max}\},
\end{equation}
where $d_{max}$ is the upper bound of geodesic distance from any node to $n_t$, and we treat it as a hyperparameter.

With the topology-aware candidate set for $n_t$, we then select the topology-aware anchors similar to Eq. \ref{eq:proximity-anchors} as follows:
\begin{equation}\label{eq:topology-anchors}
\mathcal A_T(n_t)
=\argmax_{\substack{\mathcal I\subseteq\mathcal C_T(n_t), |\mathcal I|=k_T}}
\sum_{i\in \mathcal I}\operatorname{sim}(\bm z_t,\bm z_i),
\end{equation}
where $k_T$ is the number of topology-aware anchors.

\begin{algorithm}[tb]
    \caption{Optimization Algorithm of \method{}}
    \label{alg:algorithm}
    \textbf{Requires}: a $\mathcal{G} = \langle \mathcal V, \mathcal E, \bm X\rangle$, the set of nodes with labels $\mathcal S$, and noisy labels $\mathcal Y=\{y_i\mid i\in \mathcal S\}$\\
    \textbf{Ensures}: the optimized model $\mathcal M=f_\theta\circ g_\phi$.

    \begin{algorithmic}[1] 
    \STATE Initialize the parameters of model $\mathcal M$;\\
    \STATE Compute the candidate sets $\mathcal C_P(n_t)$ and $\mathcal C_T(n_t)$ for $t\in \mathcal S$ according to Eq. \ref{eq:candidate-p} and Eq. \ref{eq:candidate-t};
    \REPEAT
    \STATE Compute the latent representations of each node, \emph{i.e.}, $\bm z_i,i\in\mathcal V$ using the encoder $f_\theta$;
    \STATE Select the proximity-aware anchors, \emph{i.e.}, $\mathcal A_P(n_t)$, from $\mathcal C_P(n_t)$ for $t\in \mathcal S$ according to Eq. \ref{eq:proximity-anchors};
    \STATE Select the topology-aware anchors, \emph{i.e.}, $\mathcal A_T(n_t)$, from $\mathcal C_T(n_t)$ for $t\in \mathcal S$ according to Eq. \ref{eq:topology-anchors};
    \STATE Obtain the set of anchors using Eq. \ref{eq:anchor-union};
    \STATE Compute the semantic homogeneity of each labeled node, \emph{i.e.}, $\mathcal H(n_t;\mathcal A(n_t);\tau)$, using Eq. \ref{eq:semantic-homo};
    \STATE Compute the class probability distributions of each labeled nodes, \emph{i.e.}, $\bm p_i,i\in\mathcal S$, using $g_\phi$;
    \STATE Compute the loss function in Eq. \ref{eq:final-loss}.
    \STATE Back-propagate through the model $\mathcal M$ and update the model's parameters.
    \UNTIL{convergence}
    \end{algorithmic}
\end{algorithm}

\subsection{Relation-Informed Dynamic Optimization}
After the proximity-aware and topology-aware anchors have been selected, we can now leverage the semantic homogeneity of the target node $n_t$ and the anchors to determine the reliability of the label $y_t$. A correctly labeled node will have many semantically similar nodes at both the global categorical level (proximity-aware anchors) and local sub-categorical level (topology-aware anchors). Conversely, for an incorrectly labeled node, the anchor selection processes will include many less similar nodes, as the label $y_t$ does not reflect the node’s true semantics.

Specifically, we first construct the set of anchor nodes relative to $n_t$ as follows:
\begin{equation}\label{eq:anchor-union}
\mathcal A(n_t) = \mathcal A_P(n_t)\cup \mathcal A_T(n_t).
\end{equation}
Subsequently, we measure the semantic homogeneity of node $n_t$ and the set of anchors according to the relations of the target node and each anchor, \emph{i.e.},
\begin{equation}\label{eq:semantic-homo}
\mathcal H(n_t;\mathcal A(n_t);\tau) = \left ( \frac{1}{|\mathcal A(n_t)|} \sum_{i\in \mathcal A(n_t)} \operatorname{sim}(\bm z_t, \bm z_i) \right )^{1/\tau},
\end{equation}
where $\tau$ is the temperature for scaling. Finally, we use the semantic homogeneity as a measurement of reliability to reweight the vanilla cross-entropy objective:
\begin{equation}\label{eq:final-loss}
\mathcal L = \frac{1}{|\mathcal S|}\sum_{i\in \mathcal S} \mathcal H(n_t;\mathcal A(n_t);\tau)\cdot \operatorname{CE}(\bm p_i,y_i),
\end{equation}
where $\operatorname{CE}(\cdot,\cdot)$ denotes the cross-entropy loss.

Moreover, since anchor selection relies on latent representations encoded by $f_\theta$, we design a dynamic optimization process that reevaluates the reliability of each labeled node during each epoch. In this way, re-evaluating reliability leads to improved supervision signals (as defined in Eq. \ref{eq:final-loss}); better supervision leads to more accurate node representations, which in turn enables better anchor selection—ultimately enhancing the model's optimization process. The overall algorithm is presented in Algorithm \ref{alg:algorithm}
\begin{table*}[t]
\centering
\resizebox{\textwidth}{!}{%
\def\arraystretch{1.15}
\begin{tabular}{lcccccccccccccccccc}
\Xhline{1.2pt}
\rowcolor{tableheadcolor} Dataset & \multicolumn{3}{c}{Cora} & \multicolumn{3}{c}{CiteSeer} & \multicolumn{3}{c}{PubMed} & \multicolumn{3}{c}{DBLP} & \multicolumn{3}{c}{A-Photo} & \multicolumn{3}{c}{Flickr} \\
\Xhline{0.8pt}
\rowcolor{tableheadcolor} Noise Type & pair & uniform & asym. & pair & uniform & asym. & pair & uniform & asym. & pair & uniform & asym. & pair & uniform & asym. & pair & uniform & asym. \\
\Xhline{1.0pt}
GCN        & 65.36 & 71.06 & 69.72 & 53.66 & 55.05 & 56.47 & 62.91 & 66.53 & 66.53 & 62.56 & 69.66 & 66.60 & 79.26 & 84.86 & 82.23 & \underline{45.86} & \underline{49.77} & \underline{49.54} \\
\rowcolor{tablerowcolor} S-model    & 65.59 & 71.05 & 70.04 & 53.41 & 54.68 & 56.28 & 63.16 & 66.52 & 66.89 & 62.60 & 69.68 & 66.37 & 80.06 & 85.32 & 84.22 & 45.47 & 49.42 & 49.27 \\
Coteaching & 56.87 & 61.22 & 62.09 & 48.34 & 46.75 & 47.47 & 60.45 & 64.38 & 61.76 & 56.98 & 64.60 & 61.99 & 74.56 & 82.78 & 77.84 & 41.55 & 43.53 & 44.19 \\
\rowcolor{tablerowcolor} JoCoR      & 64.52 & 71.16 & 70.84 & 56.30 & 57.14 & 59.05 & 55.88 & 59.44 & 57.27 & 62.39 & 68.60 & 66.33 & 68.48 & 75.22 & 79.51 & 41.20 & 43.61 & 44.17 \\
APL        & 65.97 & 72.37 & 70.67 & 53.48 & 54.61 & 56.02 & 64.28 & 67.56 & 67.27 & 64.25 & 69.37 & 66.81 & \underline{82.39} & \underline{86.17} & \underline{85.27} & 40.71 & 45.64 & 43.44 \\
\rowcolor{tablerowcolor} SCE        & 66.97 & 71.15 & 70.17 & 54.61 & 57.00 & 57.19 & 63.78 & 67.38 & 69.02 & 63.86 & 70.18 & 67.86 & 81.57 & 85.85 & 82.83 & 23.36 & 22.69 & 24.32 \\
Forward    & 65.96 & 72.47 & 70.01 & 54.57 & 56.05 & 56.94 & 62.68 & 65.62 & 65.97 & 63.59 & 70.27 & 67.03 & 67.29 & 65.13 & 63.60 & 22.98 & 21.49 & 19.88 \\
\rowcolor{tablerowcolor} Backward   & 65.64 & 72.06 & 70.05 & 54.62 & 55.81 & 56.93 & 62.35 & 65.61 & 65.84 & 63.59 & 70.26 & 66.72 & 51.90 & 55.97 & 54.90 & 25.40 & 23.93 & 22.74 \\
NRGNN      & \underline{69.73} & \underline{74.86} & \underline{74.48} & \underline{59.28} & \underline{61.87} & 63.33 & 57.22 & 62.37 & 62.38 & \underline{69.82} & \underline{76.21} & \underline{73.83} & 71.92 & 78.30 & 72.38 & 38.79 & 42.76 & 44.52 \\
\rowcolor{tablerowcolor} RTGNN      & 63.44 & 66.33 & 66.84 & 48.39 & 51.38 & 52.57 & 59.04 & 68.30 & 66.72 & 59.26 & 66.89 & 65.59 & 70.33 & 82.08 & 83.68 & 36.50 & 43.72 & 40.85 \\
CP         & 66.12 & 71.76 & 71.09 & 54.15 & 56.72 & 57.71 & 62.67 & 68.11 & 68.17 & 65.80 & 70.49 & 68.30 & 77.45 & 83.52 & 82.02 & 40.74 & 43.13 & 42.54 \\
\rowcolor{tablerowcolor} CLNode     & 64.55 & 68.24 & 67.61 & 53.19 & 55.48 & 55.24 & 62.25 & 64.56 & 66.20 & 62.59 & 65.31 & 62.45 & 79.12 & 85.46 & 82.57 & 45.06 & 48.06 & 47.45 \\
PIGNN      & 65.04 & 70.43 & 69.03 & 55.97 & 60.71 & 59.43 & \underline{66.75} & \underline{69.02} & 70.50 & 69.31 & 71.98 & 70.76 & 79.43 & 84.90 & 83.75 & 44.61 & 49.38 & 49.06 \\
\rowcolor{tablerowcolor} DGNN       & 58.49 & 56.15 & 57.16 & 46.76 & 48.99 & 49.39 & 58.36 & 63.29 & 63.42 & 61.99 & 63.47 & 61.45 & 51.09 & 54.28 & 56.66 & 15.64 & 19.24 & 15.80 \\
RNCGLN     & 68.48 & 73.29 & 74.35 & 54.16 & 58.56 & \underline{66.12} & 63.70 & 68.50 & \underline{70.90} & 64.35 & 63.65 & 65.54 & 64.79 & 72.80 & 74.37 & 22.95 & 23.99 & 24.38 \\
\rowcolor{tablerowcolor} UnionNET   & 66.29 & 72.83 & 71.02 & 58.15 & 58.87 & 60.85 & 62.77 & 65.93 & 65.60 & 62.92 & 70.33 & 66.97 & 30.89 & 32.29 & 27.53 & 20.75 & 21.19 & 22.41 \\
CGNN       & 63.27 & 69.57 & 69.01 & 51.38 & 52.31 & 51.09 & 56.01 & 57.73 & 59.76 & 58.06 & 63.26 & 58.70 & 45.70 & 47.26 & 49.43 & 11.33 & 11.71 & 11.73 \\
\rowcolor{tablerowcolor} CR-GNN     & 65.37 & 72.03 & 70.40 & 52.45 & 55.87 & 53.92 & 62.91 & 66.42 & 69.48 & 63.79 & 70.23 & 66.94 & 37.53 & 35.24 & 31.77 & 29.38 & 31.09 & 30.19 \\
\Xhline{1.0pt}
\rowcolor{ourrowname} \textbf{\method{}}     & \textbf{74.84} & \textbf{80.40} & \textbf{79.40} & \textbf{66.49} & \textbf{71.06} & \textbf{71.20} & \textbf{69.06} & \textbf{71.75} & \textbf{72.45} & \textbf{76.57} & \textbf{79.06} & \textbf{79.34} & \textbf{83.20} & \textbf{89.19} & \textbf{87.86} & \textbf{48.39} & \textbf{53.84} & \textbf{53.55}\\
\Xhline{1.2pt}
\end{tabular}%
}
\caption{The classification accuracy on six benchmark datasets. The best results are shown in \textbf{bold} and second-best \underline{underline}.}
\label{tab:main_results}
\vspace{-7mm}
\end{table*}

\section{Theoretical Analysis}
In this section, we aim to provide a theoretical understanding of the intuition behind our proposed \method{}. Before going into the details, for any node $i\in\mathcal{S}$, we denote its ground-truth label as $\hat{y}_{i}$ and represent the corresponding input feature vector as $\bm{x}_{i}\in\mathbb R^{1\times d_{in}}$. Moreover, we assume that each input-label pair $\hat{\xi}_{i}\triangleq(\bm{x}_{i},\hat{y}_{i})$ is independently drawn from an underlying distribution $\mathcal{P}$.

Generally speaking,  the primary goal of our \method{} is to learn a probability function $\bm{p}_i(\bm{w};\bm{x}_{i})$ such that we can achieve accurate node classification, where $\bm{w}\triangleq(\theta,\phi)$ denotes all model parameters. To this aim, we often consider the following stochastic optimization problem:
\begin{equation}\label{equ:sto_problem}
\min_{\bm{w}}F(\bm{w})\triangleq\mathbb{E}_{\hat{\xi}_{i}=(\bm{x}_{i},\hat{y}_{i})\sim\mathcal{P}}\left[\operatorname{CE}\big(\bm{p}_i(\bm{w};\bm{x}_{i}),\hat{y}_{i}\big)\right].
\end{equation} 

However, in many real-world scenarios, due to factors such as human annotation error and improper data collection processes, the observed label $y_{i}$ is often corrupted causing the obtained input-label pair $\xi_{i}\triangleq(\x_{i},y_{i})$ to follow a different distribution, denoted as $\mathcal{Q}$. 

Luckily, according to the change-of-measure rules~\citep{liu2015classification} or the Radon–Nikodym theorem~\citep{durrett2019probability}, we can rewrite the objective $F(\bm{w})$ in Eq. \ref{equ:sto_problem} as an expectation over the distribution $\mathcal{Q}$, that is to say,
\begin{equation}\label{equ:sto_problem_change}
    \begin{aligned}
        &F(\bm{w})\triangleq\mathbb{E}_{\hat{\xi}_{i}=(\bm{x}_{i},\hat{y}_{i})\sim\mathcal{P}}\left[\operatorname{CE}\big(\bm{p}_i(\bm{w};\bm{x}_{i}),\hat{y}_{i}\big)\right]\\
        &=\mathbb{E}_{\xi_{i}=(\bm{x}_{i},y_{i})\sim\mathcal{Q}}\left[\frac{\text{Pr}\big((\bm{x}_{i},\hat{y}_{i})|\mathcal{P}\big)}{\text{Pr}\big((\bm{x}_{i},y_{i})|\mathcal{Q}\big)}\cdot\operatorname{CE}\big(\bm{p}_i(\bm{w};\bm{x}_{i}),y_{i}\big)\right]\\
        &=\mathbb{E}_{\xi_{i}=(\bm{x}_{i},y_{i})\sim\mathcal{Q}}\left[\alpha(\bm{x}_{i},y_i,\hat{y}_{i})\cdot\operatorname{CE}\big(\bm{p}_i(\bm{w};\bm{x}_{i}),y_{i}\big)\right],
    \end{aligned}
\end{equation} where $\alpha(\bm{x}_{i},y_i,\hat{y}_{i})\triangleq\frac{\text{Pr}((\bm{x}_{i},\hat{y}_{i})|\mathcal{P})}{\text{Pr}((\bm{x}_{i},y_{i})|\mathcal{Q})}$ is the ratio of distribution drift and the symbol $\text{Pr}(\cdot|\mathcal{P})$ or $\text{Pr}(\cdot|\mathcal{Q})$ represents the probability under the distribution $\mathcal{P}$ or $\mathcal{Q}$.

Inspired by the results of Eq. \ref{equ:sto_problem_change}, our proposed \method{} algorithm adopts a weighted cross-entropy loss (Eq. \ref{eq:final-loss}) to learn the model parameter $\bm{w}$. More importantly, for each node $i\in\mathcal{S}$, \method{} introduces a novel semantic homogeneity score $\mathcal{H}(n_i;\mathcal A(n_i);\tau)$ to approximate the unknown ratio of distribution drift $\alpha(\bm{x}_{i},\hat{y}_{i},y_i)$. In sharp contrast to previous studies, this homogeneity score fully exploits the rich relational information naturally embedded in the graph topology. Building on these findings, we next investigate the convergence of our \method{} algorithm. 

Before that, we first make some standard assumptions about the objective function $F(\bm{w})$, that is,
\begin{assumption}[Smoothness]\label{ass1}
The objective function $F(\bm{w})$ is $L$-smooth, that is, $F(\bm{w})$ is differentiable and there exists a constant $L>0$ such that 
 \begin{equation*}
  |\nabla F(\bm{w})- \nabla F(\bm{u})|\le L\|\bm{w}-\bm{u}\|_{2},
 \end{equation*} where the symbol $\|\cdot\|_{2}$ denotes $l_{2}$ norm.
\end{assumption}
\begin{assumption}[Polyak-\L ojasiewicz Condition]\label{ass2} There exists a constant $\mu>0$ such that
\begin{equation*}
    2\mu\big(F(\bm{w})-F(\bm{w}^{*})\big)\le\|\nabla F(\bm{w})\|_{2}^{2},
\end{equation*} where $\bm{w}^{*}\triangleq\mathop{\arg\min}_{\mathbf{w}}F(\bm{w})$.
\end{assumption}
\begin{remark}
It is important to highlight that the Polyak-\L ojasiewicz condition has been theoretically and empirically verified during the training of deep neural networks~\citep{li2018algorithmic,yue2023lower}.
\end{remark}
\begin{assumption}[Boundedness]\label{ass3}
For any node $i\in\mathcal{S}$, the gradient $\nabla_{\bm{w}}\operatorname{CE}\big(\bm{p}_i(\bm{w};\bm{x}_{i}),y_{i}\big)$ is bounded, that is, $\|\nabla_{\bm{w}}\operatorname{CE}\big(\bm{p}_i(\bm{w};\bm{x}_{i}),y_{i}\big)\|_{2}\le G$.
\end{assumption}
Furthermore, we suppose, at every outer iteration $k$ of Algorithm~\ref{alg:algorithm}, the model parameters $\bm{w}$ are updated using standard gradient descent, \emph{i.e.}, $\bm{w}_{k+1}\triangleq\bm{w}_{k}-\eta\nabla\mathcal{L}(\bm{w}_{k})$, where $\eta$ denotes the step size and $\mathcal{L}(\bm{w})$ represents the loss function utilized by our \method{} method (See Eq. \ref{eq:final-loss}). With all these preparations, we can have the following theorem:
\begin{theorem}[Proof is deferred to Appendix]\label{thm1} Under Assumption~\ref{ass1}-\ref{ass3}, if the semantic homogeneity $\mathcal{H}(n_i;\mathcal A(n_i);\tau)$ of any node $i\in\mathcal{S}$ is a $(\beta,\epsilon)$-approximation to the $\alpha(\bm{x}_{i},\hat{y}_{i},y_i)$, i.e., $|\mathcal{H}(n_i;\mathcal A(n_i);\tau)-\beta\cdot\alpha(\bm{x}_{i},\hat{y}_{i},y_i)|\le\epsilon$, the step size $\eta\triangleq\mathcal{O}\big(\frac{1}{\beta L}\big)$ and the size of labeled nodes $|S|\triangleq\mathcal{O}(\frac{1}{\epsilon})$, after $k$ iterations, the final model parameters $\bm{w}_{k+1}$ yielded by our proposed \method{} satisfies that: 
\begin{equation*}
\mathbb{E}\big(F(\bm{w}_{k+1})-F(\bm{w}^{*})\big)\le\mathcal{O}\left(\left(1-\frac{\mu}{3L}\right)^{k}+\frac{\beta^{2}G^{2}}{\mu}\epsilon\right),
\end{equation*} where $\bm{w}^{*}$ is the optimal parameters setup.
\end{theorem}
\begin{remark}
Theorem~\ref{thm1} implies that when the relational information embedded in the graph is sufficiently rich to estimate the unknown ratio of distribution drift  $\alpha(\bm{x}_{i},\hat{y}_{i},y_i)$, if we set $\eta\triangleq\mathcal{O}\big(\frac{1}{\beta L}\big)$ and $|S|\triangleq\mathcal{O}(\frac{1}{\epsilon})$, after $\mathcal{O}(\frac{L}{\mu}\log(\frac{1}{\epsilon}))$ iterations, our proposed \method{} method can achieve an $\epsilon$-error approximation to the optimal model parameters $\bm{w}^{*}$.
\end{remark}

\section{Experiments}
\subsection{Experimental Setup}
\noindent\textbf{Datasets.}
Our evaluation is conducted on six widely used graph benchmark datasets across various domains, including  Cora \cite{mccallum2000automating}, CiteSeer \cite{giles1998citeseer}, PubMed \cite{sen2008collective}, DBLP \cite{pan2016tri}, A-Photo \cite{shchur2018pitfalls}, and Flickr \cite{yang2020scaling}. 
Both homophilic and heterophilic graphs are included. 

\noindent\textbf{Compared Baselines.}
The proposed method is compared against a comprehensive suite of state-of-the-art baselines, categorized as follows.
\textit{(1) Vanilla GNN}, including GCN \cite{kipf2016semi}. 
\textit{(2) Learning with Label Noise (LLN) methods}, including S-model \cite{goldberger2017training}, Coteaching \cite{han2018co}, JoCoR \cite{wei2020combating}, APL \cite{ma2020normalized}, SCE \cite{wang2019symmetric}, Forward \cite{patrini2017making}, and Backward \cite{patrini2017making}.
\textit{(3) Graph Neural Networks under Label Noise (GLN) methods}, including NRGNN \cite{dai2021nrgnn}, RTGNN \cite{qian2023robust}, CP \cite{zhang2020adversarial}, CLNode \cite{wei2023clnode}, PIGNN \cite{du2021noise}, DGNN \cite{nt2019learning}, RNCGLN \cite{zhu2024robust}, UnionNET \cite{li2021unified}, CGNN \cite{yuan2023learning}, and CR-GNN \cite{li2024contrastive}. 

\noindent\textbf{Evaluation Details.} In this paper, we generally follow the evaluation setting of NoisyGL \cite{wang2024noisygl}. The labels in the training and validation sets are subject to noise, while the test set remains clean for proper evaluation. In addition to the uniform label noise, we also adopt pair noise and asymmetric noise in NoisyGL \cite{wang2024noisygl}. We adopt the noise rate of 30\% for all noise types by default and report the average classification accuracies over ten runs.

\noindent\textbf{Implementation Details.} We adopt a two-layer GCN \cite{kipf2016semi} as the graph learning model $\mathcal M$. $f_\theta$ and $g_\phi$ are both implemented with one graph convolution layer, and the hidden dimension $d$ is fixed to 64. We also adopt contrastive learning \cite{you2020graph} following prior works \cite{yuan2023learning, li2024contrastive}. For the similarity measurement, \emph{i.e.}, $\operatorname{sim}(\cdot,\cdot)$, we use cosine similarity rescaled to the interval of $[0,1]$. By default, we set the hyperparameters as follows. In proximity-aware anchor selection, we set the number of anchors $k_P$ to 15. In topology-aware anchor selection, we set the number of anchors $k_T$ to 10. The upper bound of geodesic distance $d_{max}$ is set to 4. During the computation of semantic homogeneity, we set the temperature $\tau$ to 0.04. For optimization, we use Adam optimizer \cite{kingma2014adam} with a learning rate of $1\times 10^{-2}$ and train the model for 500 epochs on an NVIDIA H800 GPU.

\subsection{Main Results}
We report the average classification accuracy of the proposed \method{} compared to baseline methods over ten runs, and the results are shown in Table \ref{tab:main_results}. According to the results, we have the following observations.
Firstly, the proposed \method{} achieves a consistent improvement in terms of accuracy over existing baselines on all six datasets under different types of label noise. Specifically, \method{} achieves 14.9\% relative improvement compared to the second best on the CiteSeer dataset with uniform label noise. The significant improvements demonstrate the robustness of \method{} against label noise.
Secondly, learning with label noise (LLN) methods designed for Euclidean data (\emph{e.g.}, JoCoR, APL, SCE) generally perform similar to or worse than vanilla GCN. One possible explanation is that they do not consider the topological information in the graphs under label noise.
Lastly, for graph neural networks under label noise (GLN) methods, such as NRGNN, RTGNN, PIGNN, while some exhibit mild improvement over vanilla GCN, their overall performance remains unsatisfactory—particularly on heterophilic graphs (\emph{e.g.}, Flickr). These methods often struggle to distinguish reliable from unreliable nodes and fail to effectively utilize the relational information embedded in the graph topology.

\begin{table}[t]
\centering
\footnotesize
\setlength{\tabcolsep}{1.5pt}
\resizebox{0.472\textwidth}{!}{%
\def\arraystretch{1.1}
\begin{tabular}{lccccccccc}
\Xhline{1.2pt}
\rowcolor{tableheadcolor} Dataset & \multicolumn{3}{c}{Cora} & \multicolumn{3}{c}{CiteSeer} & \multicolumn{3}{c}{PubMed} \\
\Xhline{1.0pt}
\rowcolor{tableheadcolor} Noise & pair & uniform & asym. & pair & uniform & asym. & pair & uniform & asym. \\

\Xhline{1.0pt}
GCN      & 65.36 & 71.06 & 69.72 & 53.66 & 55.05 & 56.47 & 62.91 & 66.53 & 66.53 \\
\rowcolor{tablerowcolor} V1 & {74.38} & {79.60} & {78.89} & {65.42} & {69.69} & {70.07} & {68.77} & {71.48} & {72.24} \\
V2 & 70.56 & 77.04 & 75.56 & 59.90 & 62.11 & 62.77 & 67.38 & 69.88 & 70.39 \\
\rowcolor{tablerowcolor} V3 & 66.07 & 71.78 & 70.40 & 54.13 & 56.59 & 56.98 & 62.87 & 66.35 & 66.93 \\
V4 & 70.85 & 75.81 & 75.56 & 61.64 & 65.96 & 66.86 & 67.93 & 70.48 & 72.06 \\
\rowcolor{tablerowcolor} V5 & 68.88 & 74.77 & 74.56 & 60.85 & 65.16 & 66.11 & 68.29 & 70.62 & 72.07 \\
\Xhline{1.0pt}
\rowcolor{ourrowname} \textbf{\method{}} & \textbf{74.84} & \textbf{80.40} & \textbf{79.40} & \textbf{66.49} & \textbf{71.06} & \textbf{71.20} & \textbf{69.06} & \textbf{71.75} & \textbf{72.45} \\
\Xhline{1.2pt}
\end{tabular}}%
\caption{Ablation studies of the proposed \method{}.}
\label{tab:ablation_study}
\vspace{-8mm}
\end{table}

\subsection{Ablation Studies}
We then investigate how the different components of \method{} affect overall performance and present the results in Table \ref{tab:ablation_study}. Specifically, we design five variants of \method{}. V1 removes the topology-aware anchors (\emph{i.e.}, $\mathcal A_T(n_t)$) from the \method{} framework. V2 removes the proximity-aware anchors (\emph{i.e.}, $\mathcal A_P(n_t)$). V3 disables temperature scaling (\emph{i.e.}, $\tau=1$) when computing semantic homogeneity scores. V4 removes the candidate sets (\emph{i.e.}, $\mathcal C_P(n_t)$ and $\mathcal C_T(n_t)$) and selects anchors from all the nodes $\mathcal V$. V5 selects from the union of both candidate sets (\emph{i.e.}, $\mathcal C_P(n_t) \cup \mathcal C_T(n_t)$). As can be seen from the results, each component of \method{} contributes to overall performance, and removing or replacing any of them with alternatives degrades performance. Temperature scaling is an important technique in \method{}, as is demonstrated by low performance (V3) when it is disabled. One explanation is that scaling enlarges the gap between clean (reliable) and noisy (unreliable) nodes, making training more effective. Additionally, selecting anchors from candidates separately (proximity-aware and topology-aware) is also important, as suggested by the low accuracy in V4 and V5. The candidate sets provide proximity and topology awareness, filtering out irrelevant nodes less meaningful for the target nodes, while separate selection ensures balanced sampling from both sets of candidates.

\begin{figure}[t]
    \centering

    \begin{subfigure}[b]{0.5\columnwidth}
        \centering
        \includegraphics[width=\linewidth]{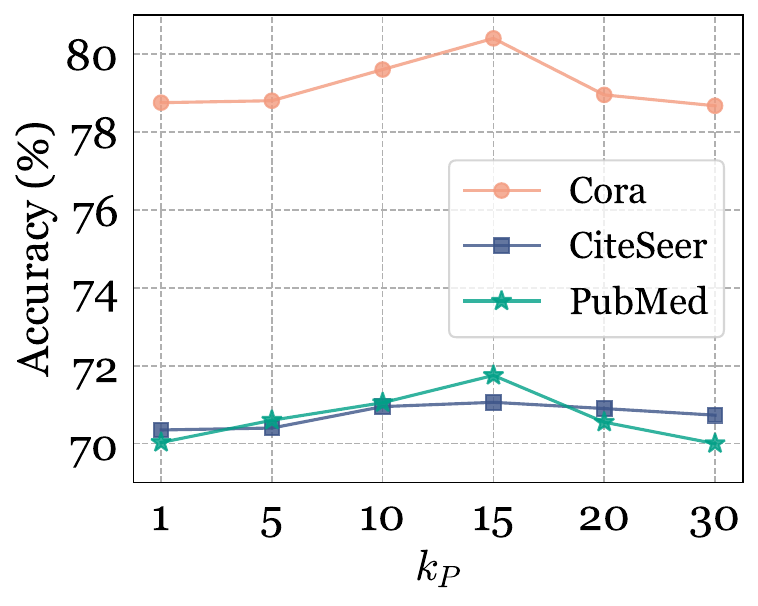}
        \vspace{-5mm}
        \label{fig:hyperparameter_k_p}
    \end{subfigure}%
    \hfill
    \begin{subfigure}[b]{0.5\columnwidth}
        \centering
        \includegraphics[width=\linewidth]{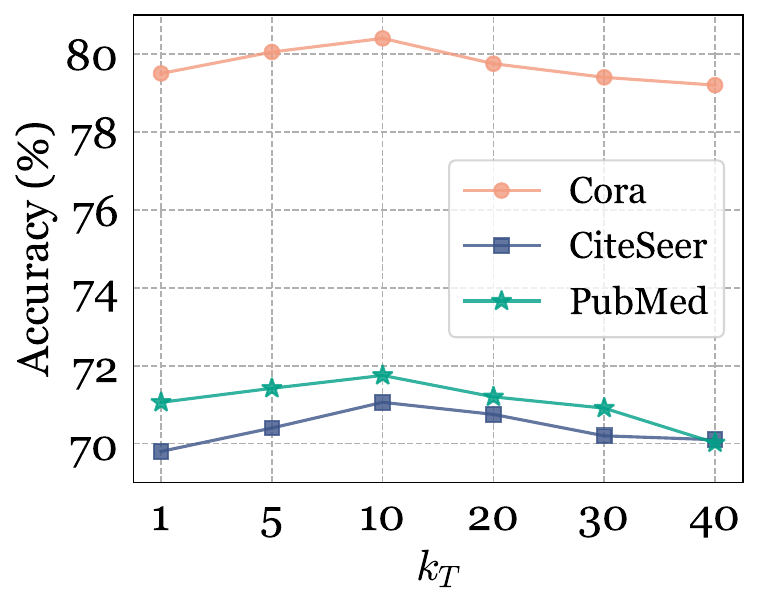}
        \vspace{-5mm}
        \label{fig:hyperparameter_k_t}
    \end{subfigure}%
    \vspace{-2mm}
    \caption{Hyperparameter analysis of \method{} on three datasets (\emph{i.e.}, Cora, CiteSeer, and PubMed). We show the accuracy under different numbers of proximity-aware anchors (\emph{i.e.}, $k_P$, left) and topology-aware anchors (\emph{i.e.}, $k_T$, right).}
    \label{fig:hyperparameter}
    \vspace{-2mm}
\end{figure}

\subsection{Hyperparameter Analysis}
We also show how the proposed \method{} performs under different hyperparameters. The results on three datasets under 30\% uniform noise are presented in Figure \ref{fig:hyperparameter}.

\noindent\textbf{Numbers of Proximity-Aware Anchors.} The prediction accuracy of \method{} under different numbers of proximity-aware anchors (\emph{i.e.}, $k_P$) is shown in Figure \ref{fig:hyperparameter} (left). As can be seen from the results, \method{} achieves best accuracy when $k_P$ is set to 15. When the number of proximity-aware anchors is too small, the model may not have enough proximity awareness for deciding the reliability of the target nodes. Conversely, when $k_P$ is too large, the selection process may include more noise from additional anchors.

\noindent\textbf{Numbers of Topology-aware Anchors.} We also present the effect of the number of topology-aware anchors (\emph{i.e.}, $k_T$) in Figure \ref{fig:hyperparameter} (right). As can be seen from the figure, our method achieves the best performance when $k_T$ is set to 10. Similar to hyperparameter $k_P$, the change of $k_T$ affects the model's topology awareness in the computation of semantic homogeneity, in addition to the balance between proximity awareness and topology awareness. A moderate value of 10 topology-aware anchors leads to the best result.

\begin{table}[t]
\centering
\footnotesize
\setlength{\tabcolsep}{3.5pt}
\resizebox{0.472\textwidth}{!}{%
\def\arraystretch{1.1}
\begin{tabular}{
  l
  *{6}{S[table-format=2.2]}
}
\Xhline{1.2pt}
\rowcolor{tableheadcolor} Dataset & {Cora} & {CiteSeer} & {PubMed} & {DBLP} & {A-Photo} & {Flickr} \\
\Xhline{1.0pt}
NRGNN    & 4.52  & 4.36 & 72.23 & 27.03 & 41.63 & 40.15 \\
\rowcolor{tablerowcolor} RTGNN    & 7.95  & 16.13  & 73.62 & 65.46 & 98.66 & 73.87 \\
UnionNET & 26.78 & 13.54  & 9.05  & 10.75  & 37.36 & 66.11 \\
\rowcolor{tablerowcolor} PIGNN    & 3.84  & 11.83  & 52.79 & 6.74 & 22.17 & 34.06 \\
\Xhline{1.0pt}
\rowcolor{ourrowname} {\method{}} & 3.66 & 4.17 & 19.40 & 12.25 & 6.84 & 12.45 \\
\Xhline{1.2pt}
\end{tabular}}%
\caption{Average training time (in seconds) over 10 runs.}
\label{tab:mean_running_time}
\vspace{-7mm}
\end{table}
\subsection{Further Analysis}
\label{sec:further_analysis}
\textbf{Model Efficiency.}
We also evaluate the efficiency of the proposed \method{} in comparison with various baselines, as shown in Table \ref{tab:mean_running_time}, which reports the training time in seconds over ten runs. As can be seen from the data, the proposed method is relatively efficient compared to baselines, exhibiting similar or smaller training time. Additionally, the proposed \method{} is significantly faster than baselines on dense graphs—specifically, A-Photo (average degree: 31.13) and Flickr (average degree: 63.30). We attribute this to our design of candidate sets (Eq. \ref{eq:candidate-p} and Eq. \ref{eq:candidate-t}), which are precomputed before training and filter out less relevant nodes from the selection of anchors.

\begin{figure}[t]
    \centering

    \begin{subfigure}[b]{0.5\columnwidth}
        \centering
        \includegraphics[width=\linewidth]{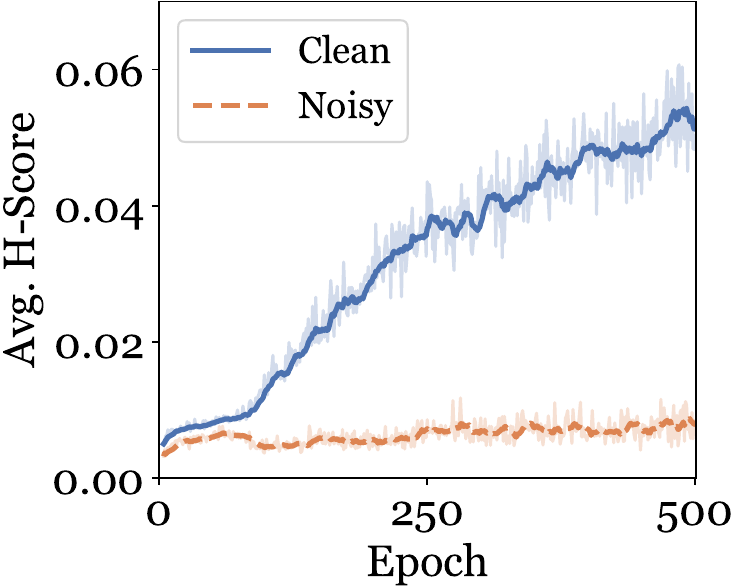}
        \vspace{-5mm}
        \caption{Cora, 30\% Uniform}
        \label{fig:hscore_cora}
    \end{subfigure}%
    \hfill
    \begin{subfigure}[b]{0.5\columnwidth}
        \centering
        \includegraphics[width=\linewidth]{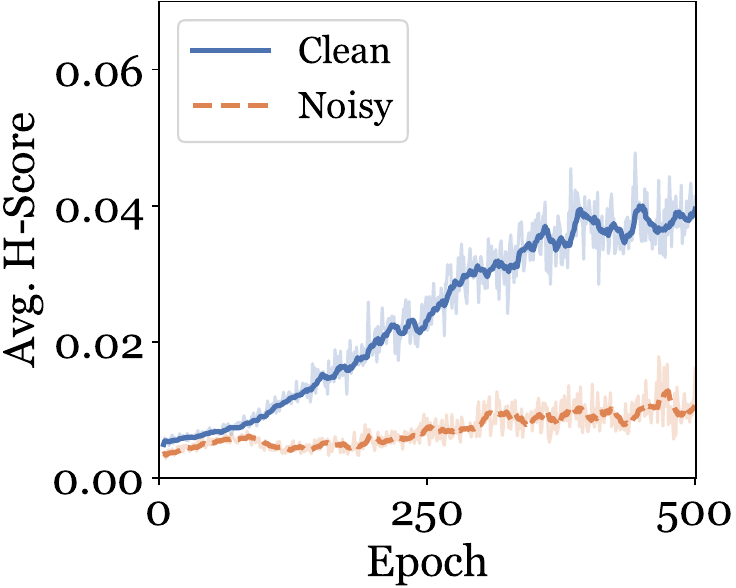}
        \vspace{-5mm}
        \caption{CiteSeer, 30\% Uniform}
        \label{fig:hscore_citeseer}
    \end{subfigure}%
    \vspace{-2mm}
    \caption{Evolution of semantic homogeneity scores of clean and noisy nodes on Cora and CiteSeer under 30\% uniform label noise. \method{} successfully separates clean from noisy nodes.}
    \vspace{-4mm}
    \label{fig:hscore}
\end{figure}

\noindent\textbf{Evolution of Semantic Homogeneity Score.} We then show that \method{} can differentiate nodes with clean labels from nodes with noisy labels dynamically during optimization. Specifically, we compute the average semantic homogeneity score in Eq. \ref{eq:semantic-homo} of nodes with clean labels and nodes with noisy labels. We present the changes in average scores of both groups of nodes during model optimization on the Cora and CiteSeer datasets under 30\% uniform label noise in Figure \ref{fig:hscore}. As shown in the figure, the proposed \method{} effectively separates clean nodes from noisy ones, and this separation is achieved dynamically throughout the optimization process. During the initial epochs of training, the average semantic homogeneity scores of clean and noisy nodes are similar. As the optimization progresses, the average scores quickly diverge, and the gap generally widens over the course of training. As the homogeneity scores are recomputed in each epoch, the weights measuring reliability dynamically change during optimization, forming a positive feedback loop: better measurement of reliability leads to improved supervision signals, and the resulting enhancement of the model enables better selection of anchors, which in turn further improves the measurement of reliability through the computed semantic homogeneity scores.

\begin{figure}[t]
    \centering

    \begin{subfigure}[ht]{0.5\columnwidth}
        \centering
        \includegraphics[width=\linewidth]{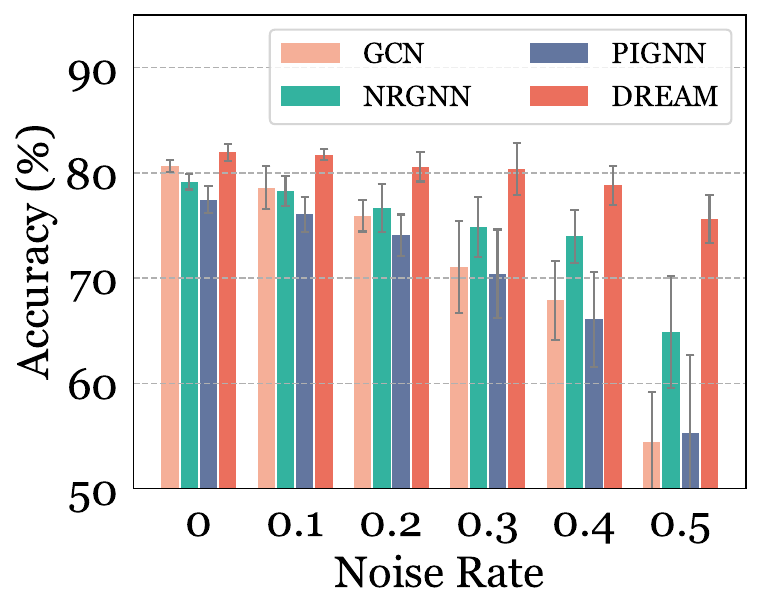}
        \vspace{-7mm}
        \caption{Cora}
        \label{fig:diff_nr_cora}
    \end{subfigure}%
    \hfill
    \begin{subfigure}[ht]{0.5\columnwidth}
        \centering
        \includegraphics[width=\linewidth]{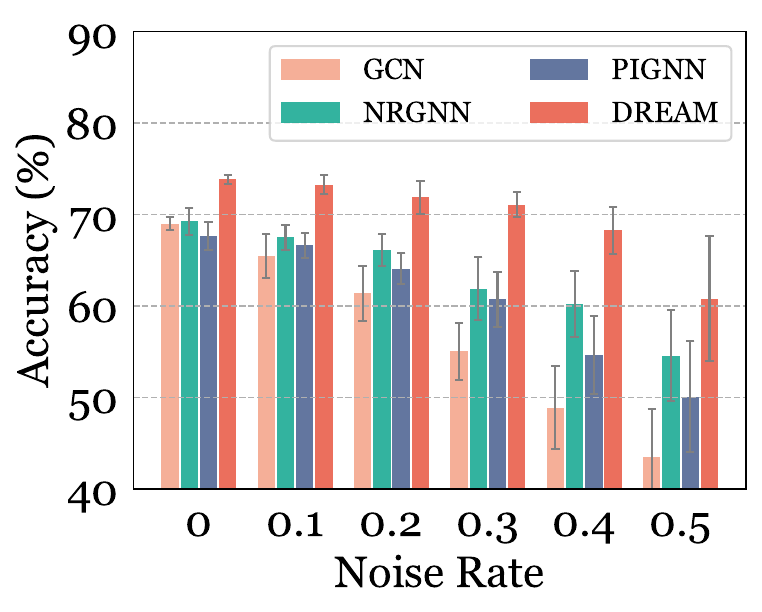}
        \vspace{-7mm}
        \caption{CiteSeer}
        \label{fig:diff_nr_citeseer}
    \end{subfigure}%
    \vspace{-2mm}
    \caption{Performance under different levels of uniform label noise on (a) Cora and (b) CiteSeer. The accuracy of the proposed \method{} degrades significantly slower than baseline methods as the noise rate increases from $0$ to $0.5$.}
    \label{fig:diff_nr}
    \vspace{-3mm}
\end{figure}

\noindent\textbf{Performance Under Different Noise Levels.}
We also show that \method{} performs consistently better than baseline methods under different noise levels. Specifically, we train our model and baselines under different noise rates ranging from $0$ to $0.5$, and compare the classification accuracy on the clean test set. The results on the Cora and CiteSeer datasets are presented in Figure \ref{fig:diff_nr}, where similar results can be observed. From the results, we can see that \method{} consistently outperforms the baselines under different noise rates on both datasets. Additionally, as the ratio of noisy labels increases, our method still achieves strong accuracy, and the performance gap between the proposed \method{} and the baselines generally widens. This suggests that our method is more robust against large noise rates and serves as a more practical solution under higher levels of label noise.

\begin{figure}[t]
    \centering

    \begin{subfigure}[ht]{0.48\columnwidth}
        \centering
        \includegraphics[width=\linewidth]{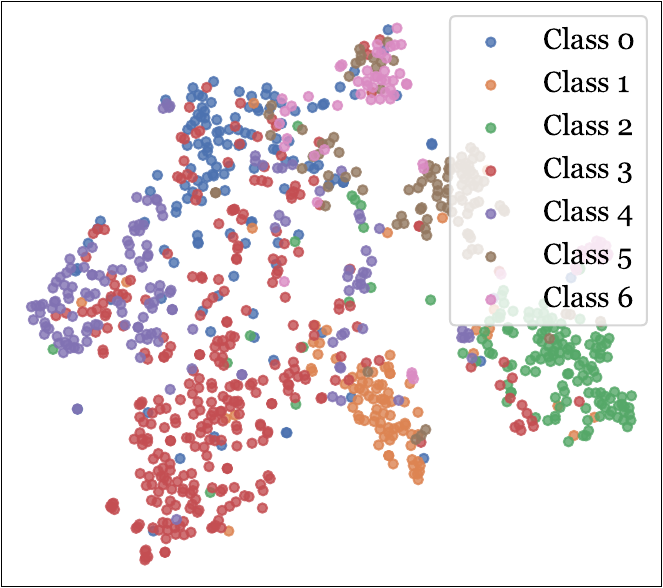}
        \vspace{-5mm}
        \caption{GCN, Cora}
        \label{fig:tsne_gcn_cora}
    \end{subfigure}%
    \hfill
    \begin{subfigure}[ht]{0.48\columnwidth}
        \centering
        \includegraphics[width=\linewidth]{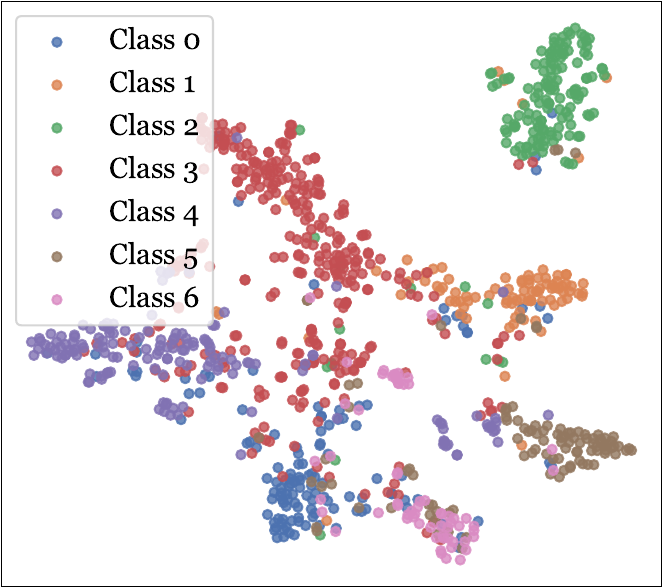}
        \vspace{-5mm}
        \caption{\method{}, Cora}
        \label{fig:tsne_dream_cora}
    \end{subfigure}%
    \vspace{-2mm}
    \caption{Learned representations under 40\% uniform label noise on Cora. The proposed \method{} yields more condensed clusters and better separates different classes compared to GCN.}
    \label{fig:tsne}
    \vspace{-6mm}
\end{figure}

\noindent\textbf{Visualization of Learned Representations.} We also provide visualization of learned node representations using t-SNE \cite{maaten2008visualizing}. Specifically, we compare the node representations generated by GCN and the proposed \method{} on the Cora dataset under 40\% uniform label noise, and the results are presented in Figure \ref{fig:tsne}. As can be seen from the figure, \method{} generates denser and more clustered node embeddings, \emph{e.g.}, the blue dots in Figure \ref{fig:tsne} (b) compared to (a). This shows that while the node embeddings learned by GCN are hampered by noisy labels, the proposed \method{} reduces the negative effect of label noise through the estimated reliability from anchor nodes.

\section{Conclusion}
This paper investigates the important problem of graph learning under label noise and proposes a novel framework named {D}ual-Standa{r}d S{e}mantic Homogeneity with Dyn{a}mic Opti{m}ization (\method{}) to enable reliable relation-informed dynamic optimization of graph learning models. In particular, we measure the reliability of each labeled node based on its relationship with a set of anchors carefully selected according to both semantic proximity and graph topology. Semantic homogeneity scores are computed to quantitatively measure the reliability and are used to reweight the optimization objective. We provide a rigorous theoretical analysis as well as extensive empirical experiments to demonstrate the effectiveness of \method{}.

\section*{Impact Statement}

This paper presents work whose goal is to advance the field of Machine
Learning. There are many potential societal consequences of our work, none
which we feel must be specifically highlighted here.


\bibliography{example_paper}
\bibliographystyle{icml2026}


\end{document}